\documentclass[letterpaper, 10 pt, conference]{ieeeconf}  % Comment this line out if you need a4paper

\IEEEoverridecommandlockouts                              % This command is only needed if 
\usepackage{graphicx}
\usepackage{float}
\usepackage{subfig}
\usepackage{multirow}
\usepackage{cite}
\usepackage{flushend}
\title{\LARGE \bf Recognition and Prediction of Surgical Gestures and Trajectories \\ Using Transformer Models in Robot-Assisted Surgery}

\author{Chang Shi$^{1*}$, Yi Zheng$^{1*}$ and Ann Majewicz Fey$^{1,2}$ % <-this % stops a space
\thanks{*The first and second authors contributed equally to this work. This work was supported by NIH \#1R01EB030125}% <-this % stops a space
\thanks{$^{1}$Walker Department of Mechanical Engineering, The University of Texas at Austin, Austin, TX 78712, USA}
\thanks{$^{2}$Department of Surgery, UT Southwestern Medical Center, Dallas, TX 75390, USA}
\thanks{Emails: chang.shi@austin.utexas.edu, yi.zheng@austin.utexas.edu}}

\begin{document}
\maketitle
\thispagestyle{empty}
\pagestyle{empty}

%%%%%%%%%%%%%%%%%%%%%%%%%%%%%%%%%%%%%%%%%%%%%%%%%%%%%%%%%%%%%%%%%%%%%%%%%%%%%%%%
\begin{abstract}
Surgical activity recognition and prediction can help provide important context in many Robot-Assisted Surgery (RAS) applications, for example, surgical progress monitoring and estimation, surgical skill evaluation, and shared control strategies during teleoperation. Transformer models were first developed for Natural Language Processing (NLP) to model word sequences and soon the method gained popularity for general sequence modeling tasks. In this paper, we propose the novel use of a Transformer model for three tasks: gesture recognition, gesture prediction, and trajectory prediction during RAS. We modify the original Transformer architecture to be able to generate the current gesture sequence, future gesture sequence, and future trajectory sequence estimations using only the current kinematic data of the surgical robot end-effectors. We evaluate our proposed models on the JHU-ISI Gesture and Skill Assessment Working Set (JIGSAWS) and use Leave-One-User-Out (LOUO) cross validation to ensure generalizability of our results. Our models achieve up to 89.3\% gesture recognition accuracy, 84.6\% gesture prediction accuracy (1 second ahead) and 2.71mm trajectory prediction error (1 second ahead). Our models are comparable to and able to outperform state-of-the-art methods while using only the kinematic data channel. This approach can enable near-real time surgical activity recognition and prediction.

\end{abstract}

%%%%%%%%%%%%%%%%%%%%%%%%%%%%%%%%%%%%%%%%%%%%%%%%%%%%%%%%%%%%%%%%%%%%%%%%%%%%%%%%
\section{INTRODUCTION}
Yang et al. defined autonomy levels for medical robots on a scale from 0 to 5, ranging from no autonomy to full automation which requires no human input~\cite{yang2017medical}. The highest level of fully automated Robotic-Assisted Surgery (RAS) is still far from reality  due to technical challenges, regulatory, legal, and ethical concerns. A more achievable short-term goal may be to envision a higher level of assistance offered by surgical robots - Robot-Enhanced Surgery (RES) - where collaborative and adaptive robot partners can leverage surgeon strengths and help overcome possible motor limitations~\cite{battaglia2021rethinking}. Recent research developments are enabling this new class of assistive surgical robots with contributions to predict surgeon intent~\cite{Qin2020}, perceive surgical states perception~\cite{qin2020fusion}, measure expertise levels~\cite{wang2018deep}, estimate procedural progress ~\cite{VanAmsterdam2020}, monitor surgeon's stress levels~\cite{zheng2022frame} and provide novel guidance through audio, augmented reality, and haptic channels~\cite{kitagawa2005effect,culjat2008pneumatic,ershad2021adaptive,Zheng2022styles}. 

The first step for these RES applications is often to perceive the current activities of surgical task and to predict the future surgical activities based on current activities~\cite{van2021gesture,gao2020,gao2021future,park2021}. Segmentation and recognition can be used to decompose surgical tasks, for example, suturing, into a sequence of surgical gestures (e.g., reaching needle, position needle), and perform surgical skill evaluation based on the executed sequence of gestures~\cite{tao2013surgical}. Padoy and Hager introduced a collaborative control method which could assist the surgeon by recognizing the completion of manual subtasks and automated the remaining ones on a da Vinci surgical robot~\cite{padoy2011human}. Moreover, the gesture recognition can also be used to trigger appropriate information displays on either the surgeon console monitor or trainer monitor.

Another important aspect is anticipating or predicting the operator's intent and robot movements. The prediction of robotic surgical instrument's trajectory can potentially contribute to preventing collision between instruments or with obstacles; therefore, enabling a method to prevent these dangerous adverse events during RAS. The prediction of surgeon's movements or the trajectory of surgeon side manipulators can help generate reference trajectories to develop haptic feedback methods  to improve surgical training outcomes.

Surgical activity recognition and prediction are both time-series sequence modeling problems. Recurrent Neural Networks (RNN) have been widely used for time-series modeling problems, for example, Gated Recurrent Units (GRU)~\cite{cho2014gru} and Long-Short-Term-Memory (LSTM) networks~\cite{hochreiter1997lstm}. These techniques were initially aimed to solve NLP problems~\cite{cho2014gru,sundermeyer2012lstm}, but they can be easily used for solving other real-world problems such as stock price prediction and human activity recognition~\cite{selvin2017stock,singh2017human}. 

Bahdanau et al. first introduced attention in machine translation where the output will focus its attention on a certain part of a sequence~\cite{Bahdanau2015}. Although the attention mechanism has been widely studied, such attention mechanisms are primarily used in conjunction with a RNN~\cite{qin2017dual}. The Transformer model has gained popularity after being published by Vaswani et al.~\cite{vaswani2017attention}. Unlike attention-based RNN, the key feature of the Transformer model is its novel attention mechanism which avoids recurrence and only relies on attention to draw global dependencies between model input and output. It replaces the recurrent layers commonly used in encoder-decoder architectures with multi-head attention. According to Vaswani et al., it is believed that Transformer can be trained significantly faster than RNN-based architectures during translation tasks. 

\textbf{Contributions:} In this paper, we propose the use of Transformer model for surgical activity recognition and prediction. Our method relies only on kinematic data from the surgical robot and has comparable if not better performance than state-of-the-art surgical activity recognition and prediction methods.

\section{BACKGROUND}
\subsection{Prior Work in Surgical Activity Recognition}
Surgical activity recognition from robot kinematic data has been studied over the last decade. With developments in machine learning techniques, especially deep learning, the methods for surgical activity recognition have evolved from Hidden Markov Models (HMMs)~\cite{tao2012sparse} and Conditional Random Fields (CRFs)~\cite{tao2013surgical} to more complex deep learning models such as LSTM neural networks~\cite{VanAmsterdam2020,dipietro2016recognizing,Yasar2020,dipietro2019segmenting}. LSTM is an appropriate tool for time-series sequence modeling due to its inherent structure to ``memorize" and ``forget" certain points within a sequence of data. 
In addition to robot kinematic data, surgical video which directly embeds surgical activity information, was also introduced to surgical activity recognition based on the development of Convolutional Neural Networks (CNN) and computer visions~\cite{lea2016segmental}. Qin et al. recently proposed Fusion-KVE which uses Temporal Convolution Networks (TCN)~\cite{lea2016tcn} and LSTM to process multiple data sources, such as kinematic data and video for surgical gesture estimation~\cite{qin2020fusion}.
\subsection{Prior Work in Surgical Activity Prediction} Time-series prediction is another popular topic in deep learning, especially LSTM. For example, LSTM has gained its popularity in stock price prediction~\cite{selvin2017stock,mehtab2020stock}, weather forecasting~\cite{salman2018single}, etc.
The use of LSTM is relatively limited in the surgical activity prediction literature. Qin et al. introduced daVinciNet which can simultaneously predict the instrument paths and surgical states in robotic-assisted surgery~\cite{Qin2020}. The daVinciNet method uses kinematics, vision and events data sequences as an input and uses an LSTM encoder-decoder model as well as a dual-stage attention mechanism to extract information from input sequence and therefore, making predictions seconds in advance~\cite{qin2017dual}. 
Gao et al. proposed a ternary prior guided variational autoencoder model for future frame prediction in robotic surgical video sequences~\cite{gao2021future}.
\subsection{Transformer Applications}
Although the Transformer model was first designed for machine translation problems, it has been studied recently in other time-series modeling problems. Wu et al. employed a Transformer-based approach to forecasting time-series data and used influenza-like illness (ILI) forecasting as a case study. They showed that the results produced by the approach were favorably comparable to the state-of-the-art~\cite{wu2020deep}. Giuliari et al. used Transformer Network and the larger Bidirectional Transformer (BERT) to predict the future trajectories of the individual people in the scene~\cite{giuliari2021transformer}. In surgical applications, Gao et al. introduced Trans-SVNet for surgical workflow analysis~\cite{gao2021trans}.

These studies give us the confidence that the Transformer model, currently the state-of-the-art in NLP tasks, could have promising performance on time-series modeling. Therefore, inspired by the recent development and validation of Transformer model, we move a step forward to using Transformer model in RAS applications, i.e., gesture recognition, gesture and trajectory prediction. 

\section{DATASET}
The JIGSAWS dataset contains three types of surgical tasks (Knot tying, Needle passing and Suturing) completed by eight subjects in a benchtop setting using a da Vinci Surgical System~\cite{Ahmidi2017,Gao2014}. For each trial, the kinematic data of the two surgeon-side manipulators (MTMs) and two patient-side manipulators (PSMs), as well as synchronized video data, are saved. For each manipulator, the time-series kinematic data includes the end-effector position (3), rotation matrix (9), linear velocity (3), angular velocity (3) and gripper angle (1), resulting in 19 features in total for each end-effector. In addition, a key distinguishing feature of the JIGSAWS dataset is annotated gestures which are synchronized with the kinematic data. The dataset specified a common vocabulary including 15 gestures (Table~\ref{tab:gestures}~\cite{Gao2014}). We also labeled the unannotated data as $``0's"$, resulting in 16 classes in gestures. We used 39 suturing trials for model evaluation, and used all 38 kinematic features of two MTMs or PSMs. 

\section{METHOD}
We formulate the surgical activity recognition and prediction tasks as supervised machine learning tasks. We take advantage of the Transformer encoder-decoder model - a model widely used in natural language processing, to process historical information all at once, aiming at better satisfying the requirement of real-time Robot-Enhanced Surgery.

Though the three tasks of gesture recognition, gesture prediction and trajectory prediction share a similar Transformer model architecture, there are slight differences in the input and output format according to their different objectives. We define an observation window with size $T_{obs}$ and a prediction window with size $T_{pred}$. The gesture recognition task would take the input of the current kinematic data ($K$) within the $t+1$ to $t+T_{obs}$ window to generate the gesture labels ($G$) for the same window. The gesture prediction task would take the input of the current kinematic data within the $t+1$ to $t+T_{obs}$, as well as the current gesture labels ($t+1$ to $t+T_{obs}$ window) estimated by the first task, to predict the future time-series gestures labels within the $t+T_{obs}+1$ to $t+T_{obs}+T_{pred}$ window. The trajectory prediction task, we will use the current kinematic data within the $t+1$ to $t+T_{obs}$ window, together with the current gesture labels within the $t+1$ to $t+T_{obs}$ window (from task 1) and the future gesture labels within the $t+T_{obs}+1$ to $t+T_{obs}+T_{pred}$ window (from task 2), to predict the future time-series end-effector trajectory ($P$) within the time window of $t+T_{obs}+1$ to $t+T_{obs}+T_{pred}$.

\begin{table}[]
\centering
\caption{Gesture Descriptions in JIGSAWS}
\label{tab:gestures}
\begin{tabular}{c|l}
\hline
Gesture ID & Description                                     \\ \hline
G1         & Reaching for needle with right hand             \\ \hline
G2         & Positioning needle                              \\ \hline
G3         & Pushing needle through tissue                   \\ \hline
G4         & Transferring needle from left to right          \\ \hline
G5         & Moving to center with needle in grip            \\ \hline
G6         & Pulling suture with left hand                   \\ \hline
G7         & Pulling suture with right hand                  \\ \hline
G8         & Orienting needle                                \\ \hline
G9         & Using right hand to help tighten suture         \\ \hline
G10        & Loosening more suture                           \\ \hline
G11        & Dropping suture at end and moving to end points \\ \hline
G12        & Reaching for needle with left hand              \\ \hline
G13        & Making C loop around right hand                 \\ \hline
G14        & Reaching for suture with right hand             \\ \hline
G15        & Pulling suture with both hands                 
\end{tabular}%
\end{table}

\begin{figure*}[ht]
    \centering
    \includegraphics[width=0.9\linewidth]{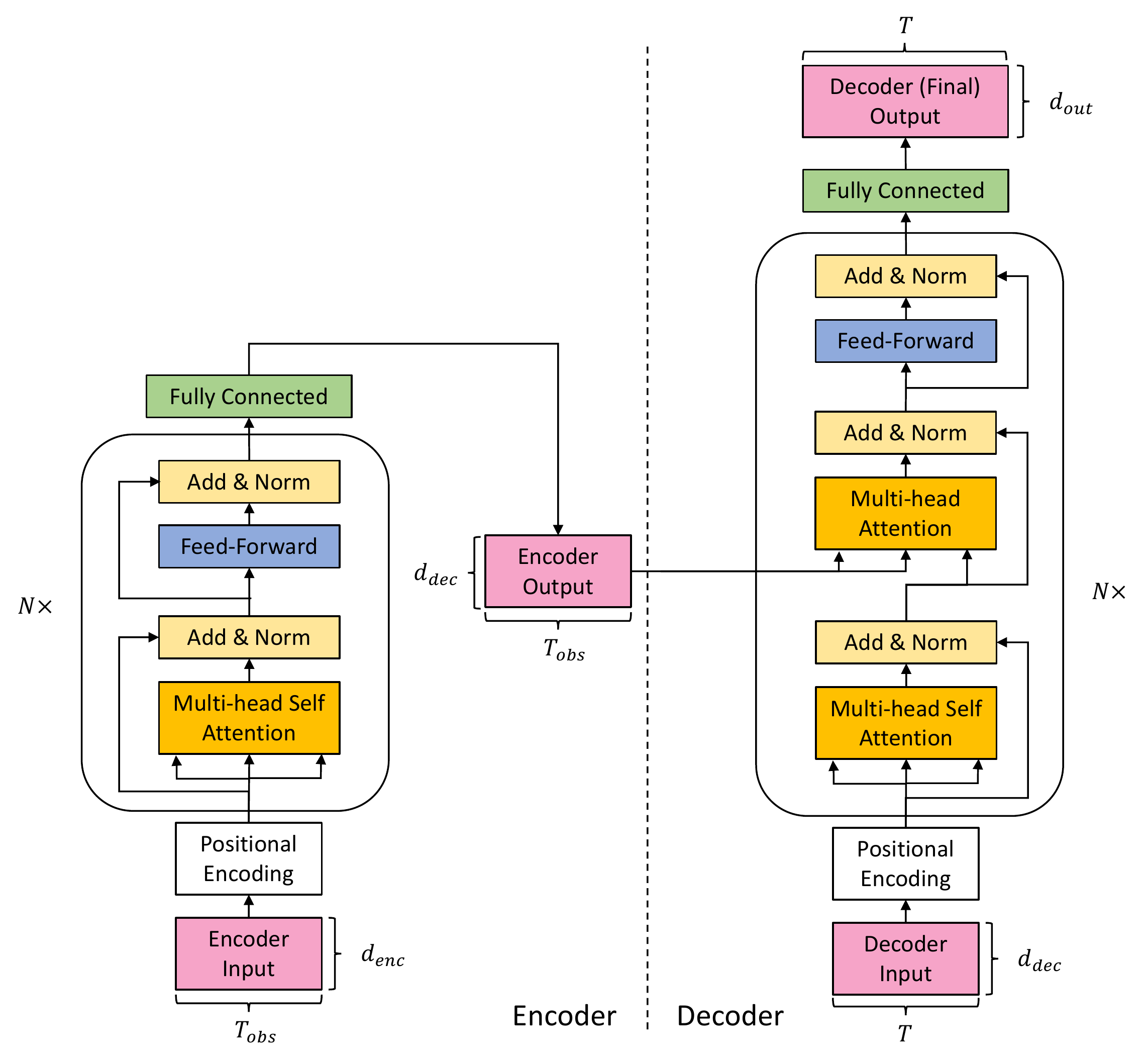}
    \caption{Architecture of the proposed Transformer model. In Encoder, $d_{enc} = 38$ during gesture recognition and prediction, $d_{enc} = 54$ during trajectory prediction. In Decoder, $T = T_{obs}$ during gesture recognition and $T = T_{pred}$ during gesture and trajectory prediction; $d_{dec} = 16$ during gesture recognition and prediction, $d_{dec} = 22$ during trajectory prediction; $d_{out} = 16$ during gesture recognition and prediction, $d_{out} = 6$ during trajectory prediction.}
    \label{fig:architecture}
\end{figure*}

\subsection{Transformer Model}

\begin{table*}[]
\centering
\caption{List of modifications based on original Transformer model~\cite{vaswani2017attention}.}
\label{tab:changes}
\begin{tabular}{ll}
\hline
Modification                 & Description                                                          \\ \hline
Removing the Embedding layer & It is designed for NLP tasks in which words are embedded to vectors. \\ \hline
Removing the Padding mask &
  \begin{tabular}[c]{@{}l@{}}It is designed for NLP tasks in which sentence lengths are different. \\ Padding mask ensures that the loss can be calculated efficiently.\end{tabular} \\ \hline
Adding Encoder output layer &
  \begin{tabular}[c]{@{}l@{}}It is a fully connected layer to map the encoder output's dimension to decoder dimension,\\  in order to calculate the attention between encoder sequence and decoder sequence.\end{tabular}
\end{tabular}%
\end{table*}

Our proposed Transformer model resembles the original Transformer architecture which consists of an Encoder and a Decoder~\cite{vaswani2017attention}. We made modifications on the original Transformer architecture based on our needs, for example, removing the embedding layers which were designed for machine translation tasks (Fig~\ref{fig:architecture} and Table~\ref{tab:changes}). For each trial, we used a sliding window (Window size: $T_{obs} = 1\,second$, stride: $S = 1\,sample$) to organize the kinematic data into frames.
\paragraph{Encoder} The Encoder consists of a positional encoding layer, a stack of $N$ identical encoder layers, and an output layer. The encoder dimension $d_{enc}$ is determined by the number of features of the encoder input sequence ($d_{enc} = 38$ for gesture recognition and prediction, $d_{enc} = 54$ for trajectory prediction). In order to make use of the order of the encoder input sequence, positional encoding is used to encode sequential information in the encoder input sequence by element-wise addition of the encoder input sequence with a positional encoding vector. Then, the resulting sequence is fed into encoder layers. Each encoder layer has two sub-layers: a multi-head self-attention mechanism, and a fully connected feed-forward network. The encoder layer is repeated for $N$ times. Finally, the data is passed through a fully connected output layer to map the data from encoder dimension $d_{enc}$ to decoder dimension $d_{dec}$. 

\paragraph{Decoder} Similarly, the Decoder consists of a positional encoding layer, a stack of $N$ identical decoder layers, and an output layer. The decoder dimension $d_{dec}$ is determined by the number of features of the decoder input sequence ($d_{dec} = 16$ for gesture recognition and prediction, $d_{dec} = 22$ for trajectory prediction). The decoder input will be discussed in the following sections. After positional encoding, the sequence is fed to decoder layers. The decoder layer has an additional multi-head attention layer between the multi-head self-attention mechanism and the fully connected feed-forward network. The added multi-head attention layer performs the attention over the output of the encoder stack. Finally, the sequence passes a fully connected layer with a dimension of $d_{out}$ to generate the results. We also employed the look-ahead masking on the decoder input sequence to ensure that the prediction of a time-series data point will only depend on previous data points.

\begin{figure}
    \centering
    \subfloat[Gesture Recognition training method.]{
    \includegraphics[width=\linewidth]{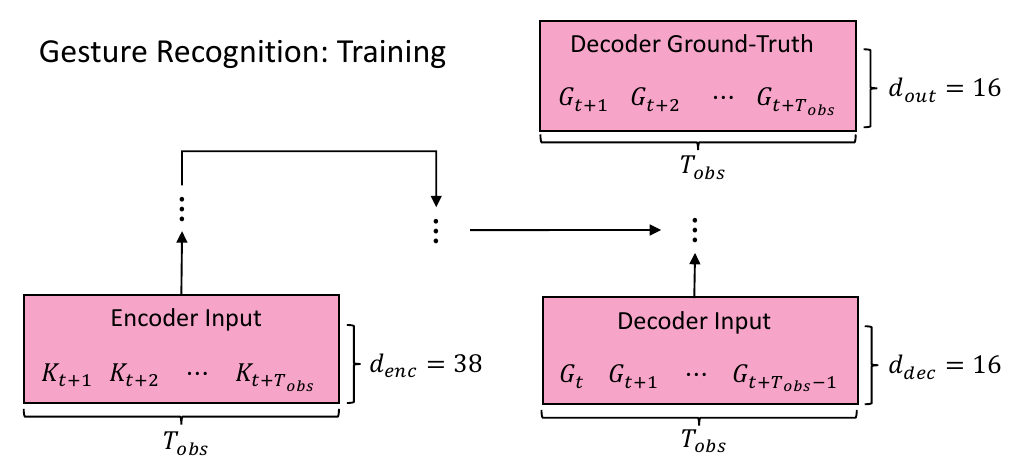}\label{fig:recog_train}}
    \\
    \subfloat[Gesture Recognition inference method.]{
    \includegraphics[width=\linewidth]{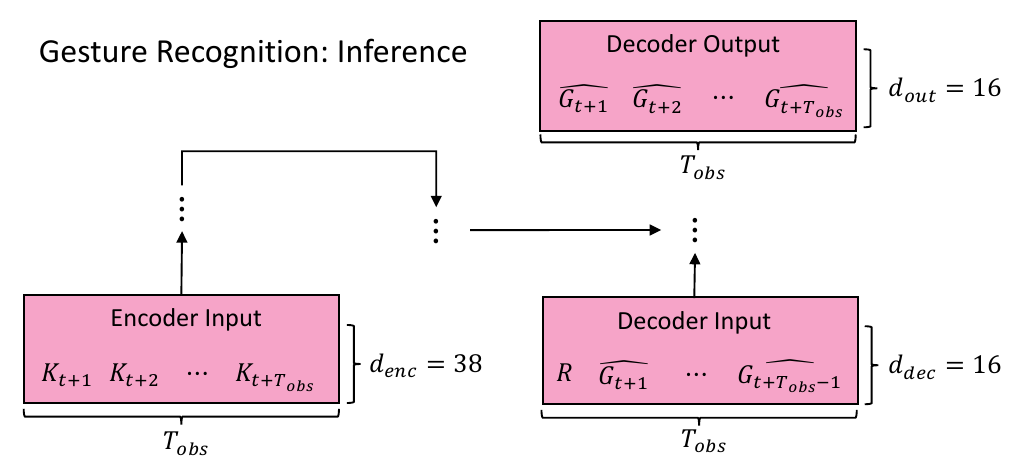}\label{fig:recog_infer}}
    \caption{Training and inference (testing) methods during Gesture Recognition. $R$ in (b) is a random vector for initializing inference. $\hat{G_i}$ in (b) is the estimated gesture value by the model. \label{fig:recog_architecture}}

\end{figure}

\begin{figure}
    \centering
    \includegraphics[width=\linewidth]{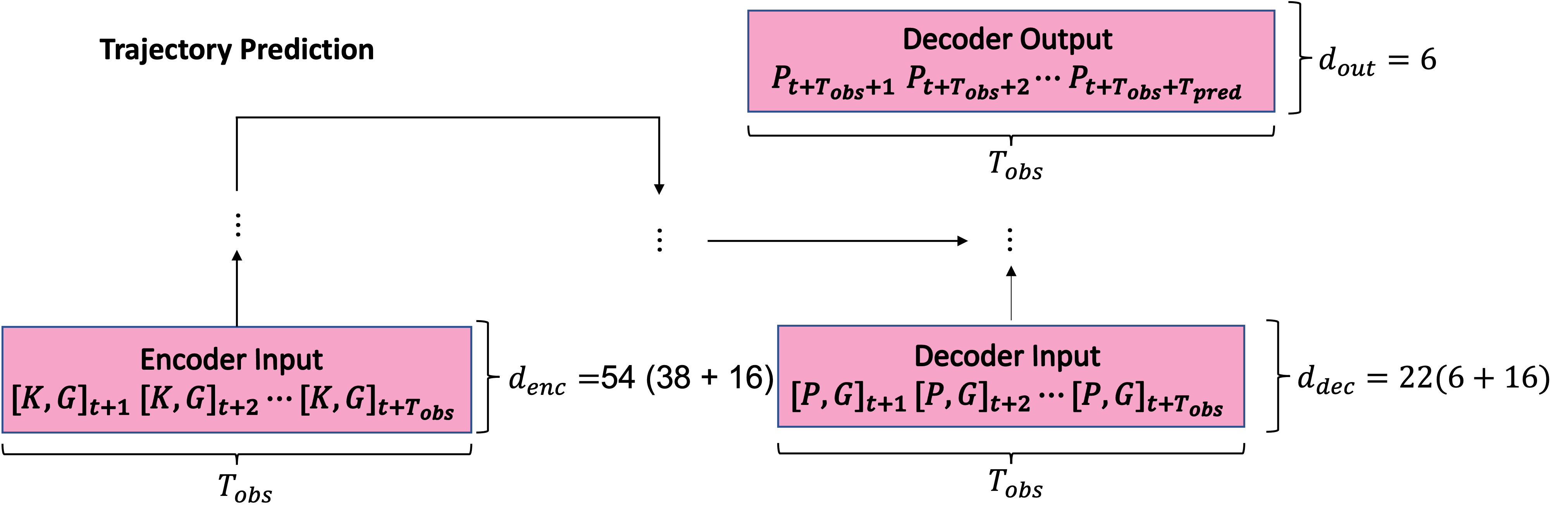}
    \caption{Training and inference methods during Trajectory Prediction.}
    \label{fig:traj_architecture}
\end{figure}

\subsection{Training and Testing}
Transformer model has three important hyperparameters that can significantly affect the model performance: Number of Encoder/Decoder Layers ($N$ in Figure~\ref{fig:architecture}), Number of Encoder Heads ($h_{enc}$), Number of Decoder Heads ($h_{dec}$). 
The hyperparameters were tuned using grid search. We shuffled the data frames of all 39 suturing trials in JIGSAWS and splitted the training/testing set by 70/30 split for grid search purposes. 
\paragraph{Gesture Recognition}
Gesture recognition can be treated as ``translating" from current kinematic data to current gestures. During training, the encoder input sequence consisted of all 38 current kinematic features of either MTMs or PSMs ($K_i$, $i = t+1, t+2, ..., t+T_{obs}$). The decoder ground-truth (output) sequence consisted of all 16 gesture classes (current gestures) of the input time steps ($G_i$, $i = t+1, t+2, ..., t+T_{obs}$). Following the teacher forcing procedure, the decoder input sequence was the shifted-right ground-truth sequence ($G_i$, $i = t, t+1, ..., t+T_{obs}-1$), as shown in Fig~\ref{fig:recog_train}. 
During testing and inference, shown in Fig~\ref{fig:recog_infer}, the first instance in decoder input was a randomized vector $R$. Then, for each inference step, the predicted value $\hat{G_i}$ from decoder output was added to the decoder input sequence recurrently for the next inference step.

\paragraph{Gesture Prediction} Similar to gesture recognition, the encoder input sequence consisted of all 38 current kinematic features of either MTMs or PSMs ($K_i$, $i = t+1, t+2, ..., t+T_{obs}$). The decoder ground-truth (output) sequence consisted of all 16 gesture classes of the future time steps ($G_i$, $i = t+T_{obs}+1, t+T_{obs}+2, ..., t+T_{obs}+T_{pred}$). The decoder input sequence consisted of the current gesture classes ($G_i$, $i = t+1, t+2, ..., t+T_{obs}$).

During testing and inference, since the inputs of Encoder and Decoder were known: current kinematic data $K$ from $t+1$ to $t+T_{obs}$ and current gesture data $G$ from $t+1$ to $t+T_{obs}$, respectively, the inference did not have any recurrence. Both inputs can be fed into the model at the same time. In both gesture recognition and gesture prediction, cumulative categorical cross-entropy loss is used for the discrepancies between the model output and the ground-truth gesture label.

\paragraph{Trajectory Prediction}
The encoder input sequence consisted of the concatenation of all 38 current kinematic features and the one-hot vector of the 16 gesture class, of PSMs ($[K_i, G_i]$, $i = t+1, t+2, ..., t+T_{obs}$). The decoder ground-truth (output) sequence consisted of 6 position dimensions $x,y,z$ of both left and right end-effectors during the future time steps ($P_i$, $i = t+T_{obs}+1, t+T_{obs}+2, ..., t+T_{obs}+T_{pred}$). The decoder input sequence consisted of the current position information ($P_i$, $i = t+1, t+2, ..., t+T_{obs}$) and the future gesture information ($G_i$, $i = t+T_{obs}+1, t+T_{obs}+2, ..., t+T_{obs}+T_{pred}$), as shown in Fig~\ref{fig:traj_architecture}. We use the cumulative $L_2$ loss between the predicted end-effector trajectory and the ground-truth trajectory, summed over the $T_{pred}$, as the trajectory loss function. To take the advantage of transformer for processing time series all at once, no recurrent inference as gesture recognition is used. Both inputs are fed into the model all at the same time. 

Similar to the original Transformer implementation, We used the Adam optimizer~\cite{kingma2014adam} with $\beta_1 = 0.9$, $\beta_2 = 0.98$ and $\epsilon = 10^{-9}$ and varied the learning rate ($lr$) over training steps~\cite{vaswani2017attention}. We used $warmup\_steps = 2000$ :
\begin{equation}
    lr = d^{-0.5}_{dec}*(steps^{-0.5}, steps*warmup\_steps^{-1.5})
\end{equation}

\section{EXPERIMENTAL EVALUATIONS}
We evaluated our gesture recognition, gesture prediction and trajectory prediction models on the JIGSAWS dataset (See Table~\ref{tab:gestures}).

To evaluate gesture recognition and gesture prediction accuracy, for each data frame, we calculated the percentage of accurately recognized or predicted time steps in the frame. Then, the accuracy was averaged across all the frames in the testing dataset. 

To evaluate the performance of end-effectors trajectory prediction, Root Mean Squared Error (RMSE) and Mean Absolute Error (MAE) were used:
\begin{equation}
RMSE=\sqrt{\frac{\sum_{i=1}^{N}\left(y^{i}-\hat{y}^{i}\right)^{2}}{N}}
\end{equation}
\begin{equation}
MAE=\frac{\sum_{i=1}^{N}\left|y^{i}-\hat{y}^{i}\right|}{N}
\end{equation} We calculated both metrics for each dimension of the Cartesian end-effector path in the endoscopic reference frame (x,y,z) and also the end-effector distance $d=\sqrt{x^2+y^2+z^2}$ from the origin (camera tip).

We adopted the Leave-One-User-Out cross validation (LOUO) to train and test the generalizability of our model. In LOUO, for each iteration, the data of $i^{th}$ subject was left out as testing set, and the rest of the data for training. Then we averaged the evaluation metrics across all the iterations which led to averaging over all subjects as testing set and reported the mean. The $batch\_size = 64$ and the model was trained for $epoch = 15$ during gesture recognition, $epoch = 40$ during gesture prediction, and $epoch = 50$ during trajectory prediction. For the final evaluation of the trajectory prediction performance, the evaluation metrics are calculated only on the time step of $T_{pred}$ without accounting for the previous prediction steps in the entire prediction window.

\section{RESULTS AND DISCUSSIONS}
In order to compare with previous studies in the literature, during gesture recognition we kept the data as its original frequency of 30Hz. During gesture and trajectory prediction, we downsampled the data to 10Hz. For each task, the model was trained and evaluated independently. 
\subsection{Gesture Recognition}
We kept the JIGSAWS data as its original frequency 30Hz. We did not run hyperparameter tuning for gesture recognition as it would take significant computational effort with the data of 30Hz. Instead, we decided to keep the model for gesture recognition in its simplest form: $N = 1$, $h_{enc} = 1$, and $h_{dec} = 1$. The encoder input sequence was 38 dimensional kinematic features with a length of current observation $T_{obs} = 1s\,(30\,samples)$. The decoder output was a 16-class gesture with a length of current observation $T_{obs} = 1s\,(30\,samples)$ of the corresponding encoder input sequence. 

\begin{figure*}[ht]
    \centering
    \includegraphics[width=\linewidth]{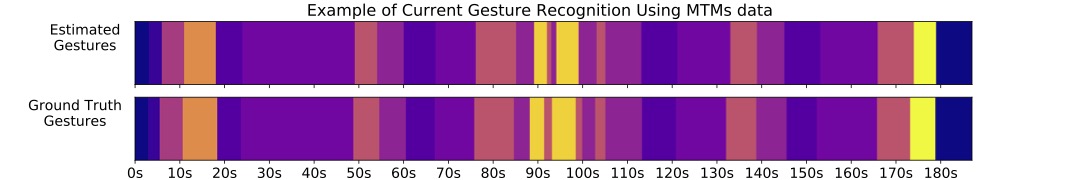}
    \caption{An example of Gesture Recognition using a random suturing trial. The top row is the estimated gestures by our proposed Gesture recognition model. The bottom row is the ground-truth labels.}
    \label{fig:recog_ex}
\end{figure*}

We used the data of MTMs and PSMs individually to test the model performance. After LOUO cross validation, the reported accuracy was promising and outperformed the state-of-the-art algorithms (89.3\% with MTMs kinematic data as encoder input; 89.2\% with PSMs kinematic data as encoder input, in Table~\ref{tab:recognition_compare}). An example of gesture recognition using the kinematic data of PSMs of a random trial as testing dataset is illustrated in Fig~\ref{fig:recog_ex}.

\begin{figure*}[ht]
    \centering
    \includegraphics[width=\linewidth]{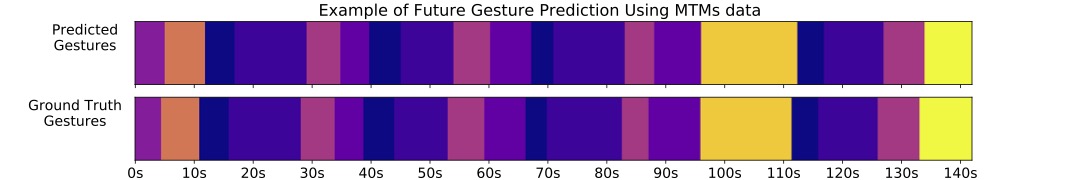}
    \caption{An example of Gesture Prediction using a random suturing trial. The top row is the predicted gestures by our proposed gesture prediction model. The bottom row is the ground-truth labels. The Decoder uses real current gestures as input.}
    \label{fig:pred_ex}
\end{figure*}

Fusion-KVE is a method which incorporates kinematic data, video and events data. However, our proposed Transformer model only uses kinematic data and outperforms Fusion-KVE in accuracy. Using fewer types of data could potentially shorten the computational time for gesture recognition and therefore, enables a near-real-time recognition manner. 
\begin{table}[]
\centering
\caption{Comparison to prior works of Gesture Recognition under LOUO cross validation. The listed models used the data of 30Hz.}
\label{tab:recognition_compare}
\begin{tabular}{lll}
\hline
                     & Data Sources  & Accuracy        \\ \hline
Fusion-KVE~\cite{qin2020fusion}           & PSMs+Video       & 86.3\%          \\
Forward LSTM~\cite{dipietro2016recognizing}         & PSMs       & 80.5\%          \\
Bidir. LSTM~\cite{dipietro2016recognizing}          & PSMs     & 83.3\%          \\
\textbf{Transformer} & \textbf{MTMs} & \textbf{89.3\%} \\
\textbf{Transformer} & \textbf{PSMs} & \textbf{89.2\%} \\
\end{tabular}%
\end{table}

\subsection{Gesture Prediction}
In order to compare our proposed model with the state-of-the-art, we also downsampled JIGSAWS data to 10Hz. The downsampled data could shorten the computational time during training, therefore, we applied hyperparameter tuning using grid search to optimize the prediction performance. After grid search, the resulting hyperparameters were: $N = 4$, $h_{enc} = 1$ and $h_{dec} = 4$. 

During training, the encoder input sequence was 38 dimensional kinematic features of current observation ($K_i$, $i = t-T_{obs}+1, t-T_{obs}+2, ..., t$,  $T_{obs} = 1s$). The decoder output sequence was the prediction of 16-class future gestures ($G_i$, $i = t+1, t+2, ..., t+T_{pred}$). It is worth noting that the decoder input sequence is the current gesture sequence ($G_i$, $i = t+1, t+2, ..., t+T_{obs}$). 

Although our proposed gesture recognition model could output a good estimation of the current gesture sequence, and it's more reasonable to use the current gesture estimation to mimic the real-world application, we decided to train and evaluate the gesture prediction model on "real" current gestures, assuming a perfect current gesture estimation during gesture recognition. It allowed us to independently evaluate the performance of gesture prediction model.

We summarized the gesture prediction model performance in Table~\ref{tab:ges_prediction_compare}. Both observation $T_{obs}$ and prediction $T_{pred}$ time lengths were 1 second. Although the reported accuracy (84.6\% with MTMs kinematic data as encoder input; 84.0\% with PSMs kinematic data as encoder input) did not significantly outperform the state of the art, we still believe our proposed gesture prediction model is promising since less data source (only kinematic data) was used in our study. One example of gesture prediction using the kinematic data of MTMs is shown in Fig~\ref{fig:pred_ex}.

\begin{table}[]
\centering
\caption{Comparison to prior works of Gesture Prediction under LOUO cross validation. The listed models used the data of 10Hz in JIGSAWS and a prediction length of 1 second.}
\label{tab:ges_prediction_compare}
\begin{tabular}{lll}
\hline
            & Data Sources & Accuracy \\ \hline
daVinciNet~\cite{Qin2020}  & PSMs+Video         & 84.3\%   \\
\textbf{Transformer} & \textbf{MTMs}  & \textbf{84.6\%}    \\
Transformer & PSMs         & 84.0\%   
\end{tabular}%
\end{table}

\begin{table*}[]
\centering
\caption{End-effector trajectory prediction performance measures with prediction window of one second ahead($T_{pred}=10$). The prediction performances are reported for the Cartesian end-effector path in the endoscopic reference frame $(x,y,z)$ and $d=\sqrt{x^2+y^2+z^2}$.}
\label{tab:traj_prediction_compare}
\begin{tabular}{lllllllllll}
\hline
            & Data Sources & Metric               & $x_1$ & $y_1$ & $z_1$ & $d_1$ & $x_2$ & $y_2$ & $z_2$ & $d_2$ \\ \hline
daVinciNet  & PSMs       & \multirow{2}{*}{RMSE} & \textbf{2.81}  & \textbf{2.42}  & \textbf{3.28}  & 4.16  & \textbf{3.8}   & 4.26  & 4.75  & 5.92  \\
Transformer & PSMs         &                      &  3.15     &   3.03    &   3.30    &   \textbf{2.93}    & 3.93     &  \textbf{4.21}    &  \textbf{4.22}   &  \textbf{4.92}   \\ \hline
daVinciNet  & PSMs       & \multirow{2}{*}{MAE} & \textbf{2.19} & \textbf{1.95}  & \textbf{2.86}  & 3.7   & \textbf{3.42}  & 3.91  & 4.31  & 5.34  \\
Transformer & PSMs         &                      &  2.86    &   2.85   &  3.00     &    \textbf{2.71}   &  3.60    &   \textbf{3.88}    &  \textbf{3.84}    &   \textbf{4.44}
\end{tabular}%
\end{table*}

\subsection{Trajectory Prediction}
For the trajectory prediction task, we downsampled JIGSAWS data to 10Hz. We used the hyperparameters: $N = 1$, $h_{enc} = 6$ and $h_{dec} = 11$. The Encoder took both the kinematic features and the 16-class gesture class as input. The Decoder took the current $x,y,z$ positions of two end-effectors and the future gestures as input. Similar to Gesture Prediction, during training and testing, we used the ground truth future gestures in decoder input, assuming a perfect gesture prediction, to evaluate the model independently.

Table~\ref{tab:traj_prediction_compare} summarizes the Transformer performance on the JIGSAWS suturing dataset with the prediction time-step of 1 second ($T_{pred}=10$). Using only the PSM data, the Transformer has better performance than daVinciNet on the right arm trajectory prediction, while slightly worse performance on the left arm. Although results are mixed, the Transformer still only uses the kinematics data to obtain as competitive results as those using both complex video and kinematic data. This would help largely reduce the computation complexity in the applications with a guarantee of accurate and real-time motion and gesture monitoring.

\section{CONCLUSIONS AND FUTURE WORK}
In this paper, we used the Transformer model, a novel deep learning model initially designed for NLP tasks, to recognize and predict the surgical activities. We modified the Transformer model architecture from the original paper according to the need of our tasks: gesture recognition, and gesture and trajectory prediction during RAS. In gesture recognition, the model took current kinematic data as its input sequence and estimated the corresponding surgical gestures (accuracy: 89.3\% using MTMs and 89.2\% using PSMs); In gesture prediction, the model took current kinematic data and current surgical gestures as its input sequences and predicted the future (1 second) surgical gestures (accuracy: 84.6\% using MTMs and 84.0\% using PSMs); In trajectory prediction, we jointly utilize the current kinematic data as well as the future gestures to predict the future (1 second) end-effector trajectory, and reached distance error as low as 2.71 mm. 

Considering that our models are purely based on the kinematic data of the end-effectors (MTMs and PSMs) of the daVinci Surgical System without the aid of visual features, the results are very much competitive. Although some studies have shown that combining kinematic data and video could improve the recognition and prediction performance, our work shows the potential to achieve similar performance with only kinematic data, which is preferred when running surgical activity recognition and prediction in a real-time manner~\cite{tao2013surgical,Zappella2013}, since vision data processing is inherently time consuming.

Though the proposed models have outperformed the state-of-the-art methods from the literature, there are still some limitations that remain unsolved. Future work would include jointly evaluating the performance of gesture and trajectory prediction models. Our gesture prediction model took the current gesture sequence as its decoder input. And our trajectory prediction model took the future gesture sequence as part of its decoder input. However, in our current implementation, we used the ground truth values of the current gestures in gesture prediction and the ground truth values of the future gestures in trajectory prediction, to train and evaluate the models independently. To test the robustness and the feasibility of the near-real-time manner of the models in real-world gesture and trajectory prediction tasks, our next step would be to use the estimated values of gesture recognition as gesture prediction decoder input, and then, use the estimated values of gesture prediction as trajectory prediction decoder input. We also plan to do more ablation studies on measuring the respective difficulty of trajectory prediction of the 16 gesture classes. This would help develop a global sense of when the RAS system should offer more motion guidance and deviation warning.

% 2. Tuning hyperparameters of gesture recognition model. We kept the data frequency as 30Hz which resulted in a slow computation, therefore, we decided to keep the model as its simplest form. To achieve the best performance, fine-tuning the hyperparameters of gesture recognition model is an important aspect of our future work.

We believe our proposed methods can contribute to implementation of robot enhanced surgical (RES) applications, therefore, augment the role of robots in assisting surgeons through modern control strategies.

%3. Including video data. Although our Gesture Prediction model could not significantly outperform the state of the art, we still believe it is possible to improve the model performance by including multiple data sources as the model input. We only had kinematic data in our study, however, daVinciNet used kinematics, video and events data. According to Qin et al., richer information regarding surgical subtasks can be extracted from multiple data sources, especially endoscope vision feature, therefore, achieving improved model performance~\cite{Qin2020}. 


%%%%%%%%%%%%%%%%%%%%%%%%%%%%%%%%%%%%%%%%%%%%%%%%%%%%%%%%%%%%%%%%%%%%%%%%%%%%%%%%

%%%%%%%%%%%%%%%%%%%%%%%%%%%%%%%%%%%%%%%%%%%%%%%%%%%%%%%%%%%%%%%%%%%%%%%%%%%%%%%%

%%%%%%%%%%%%%%%%%%%%%%%%%%%%%%%%%%%%%%%%%%%%%%%%%%%%%%%%%%%%%%%%%%%%%%%%%%%%%%%%
%%%%%%%%%%%%%%%%%%%%%%%%%%%%%%%%%%%%%%%%%%%%%%%%%%%%%%%%%%%%%%%%%%%%%%%%%%%%%%%%

\bibliographystyle{IEEEtran}
\IEEEtriggercmd{\enlargethispage{2in}}
\bibliography{transformer.bib}

\end{document}